\documentclass[11pt]{article}

\usepackage{acl}
\usepackage{times}
\usepackage{latexsym}
\usepackage{amssymb}
\usepackage{booktabs}
\usepackage{array}
\usepackage{multirow}
\usepackage{float}
\usepackage{amsmath}
\newcommand{\best}[1]{\textbf{#1}}
\usepackage[T1]{fontenc}

\usepackage[utf8]{inputenc}

\usepackage{microtype}

\usepackage{inconsolata}

\usepackage{graphicx}

\title{GeoArbiter: Verifiability-Guided Grounding for Remote-Sensing Multimodal LLMs}

\author{
  Xuechen Li \\
  University of Minnesota, Twin Cities \\
  \texttt{li003487@umn.edu}
}

\begin{document}
\maketitle
\begin{abstract}
Remote-sensing multimodal large language models (MLLMs) often assert facts that imagery cannot establish, such as a facility's identity or function. Coordinate-keyed geographic retrieval can supply this missing knowledge, improving fMoW land-use accuracy by 12.06--17.19 points across three open MLLMs. However, retrieved records can also contradict visible evidence, and we find that models frequently follow the records even when the image is decisive. We argue that source trust should therefore depend on \emph{cross-modal verifiability}: geographic records are most useful for attributes the image cannot verify and most dangerous when they dispute visually verifiable attributes. We introduce GeoArbiter, a training-free pipeline that operationalizes this principle by injecting only image-unverifiable geographic facts. Unlike arbitration prompts, which leak across attribute types and bias yes/no responses, content-level filtering preserves 84.69--87.15\% of the full-retrieval accuracy gain, reduces claim-level hallucination by 9.58--26.34\% under a source-blinded judge, and improves robustness to conflicting records across all three models. These results identify verifiability-guided content selection as a simple, effective mechanism for grounding remote-sensing MLLMs in fallible geographic knowledge.
\end{abstract}

\section{Introduction}
\label{sec:intro}

\begin{figure}[!t]
\centering
\includegraphics[width=\columnwidth]{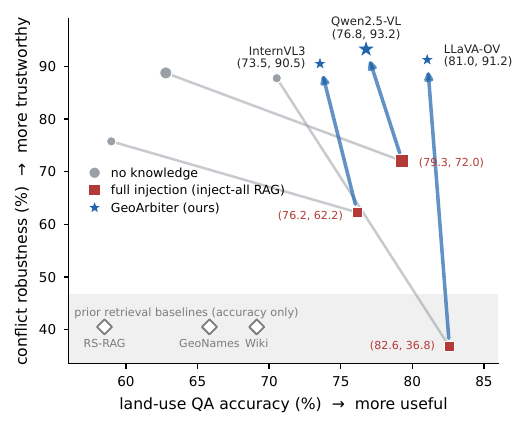}
\caption{\textbf{GeoArbiter preserves accuracy and robustness.} Full OSM injection improves land-use QA but becomes brittle to a wrong record; verifiability-stratified injection remains in the top-right across all three MLLMs.}
\label{fig:teaser}
\end{figure}

Multimodal large language models adapted to remote-sensing (RS) imagery can describe satellite scenes and answer open-ended questions fluently \citep{qwen25vl,internvl3,llavaonevision}. Yet many consequential errors are not perceptual: a model may misname a facility, invent its institutional role, or fabricate administrative context that pixels cannot establish. We call these \emph{knowledge hallucinations}, distinguishing them from errors about visible content \citep{pope,woodpecker}. They persist across RS model families \citep{rshallu,radar} and undermine applications in which an incorrect facility identity can be more costly than a missed object.

Geographic retrieval appears to offer a natural remedy. Every RS image has coordinates, which provide an exact key into resources such as OpenStreetMap (OSM; \citealp{openstreetmap}), GeoNames, and land-cover maps. Prior work, however, mainly uses these resources during training \citep{skyscript,lhrsbot,osmda,geolink,fmowmm}. Inference-time RS retrieval instead relies on visual similarity to landmark descriptions \citep{rsrag}, which poorly covers ordinary schools, substations, and depots; coordinate-keyed text systems \citep{geollm,spatialrag,georag} do not face conflicts between records and an image.

Such conflicts make grounding a source-selection problem. Geographic databases are incomplete, stale, and potentially corrupted \citep{cloudweb}; a record can supply an invisible function but can also claim that a visibly absent runway exists. Prior conflict studies examine perception, context, and parametric memory \citep{mmkcbench,insightoversight,segsub,seeingoverrides,visiondefault}, typically through synthetic contradictions and instruction- or decoding-level remedies \citep{insightoversight,cad}. They do not ask which \emph{attributes} each modality is qualified to resolve.

\begin{figure*}[t]
\centering
\includegraphics[width=\textwidth]{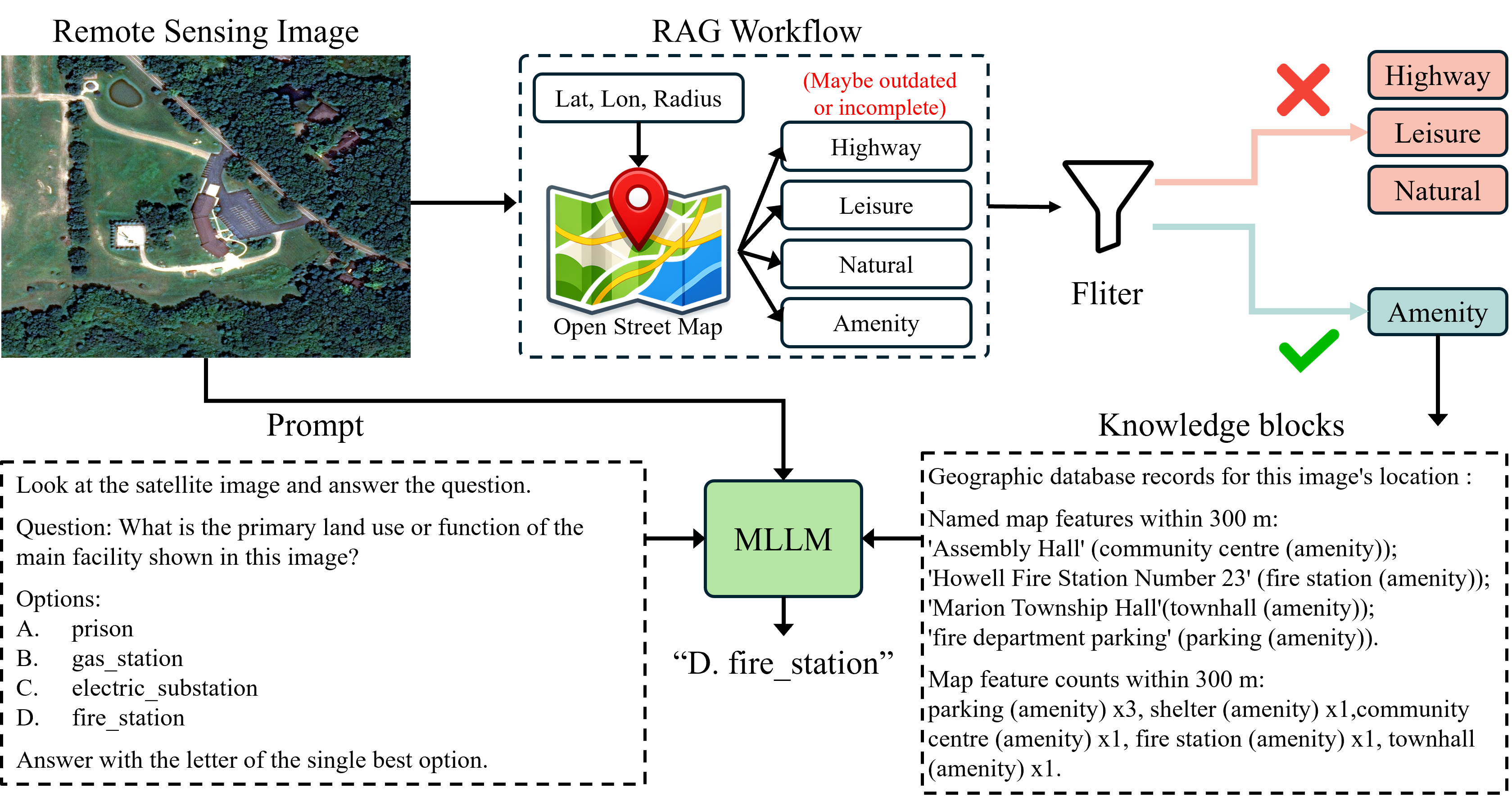}
\caption{\textbf{GeoArbiter pipeline.} Image coordinates retrieve nearby OSM features, which may be stale or incomplete. A deterministic verifiability filter withholds physical-structure keys that the image can check (red) and textualizes image-unverifiable functional records (blue) into a knowledge block. The frozen MLLM answers from the image, question, and filtered knowledge.}
\label{fig:pipeline}
\end{figure*}

We propose \emph{cross-modal verifiability} as that criterion: records should dominate for attributes that an image cannot verify (e.g., function or name), whereas the image should dominate when a record disputes visible structure. GeoArbiter enforces this policy in the retrieved content rather than asking a frozen model to follow it. A key-level filter retains function and identity records and withholds physical-structure records before prompting (Figure~\ref{fig:pipeline}).

We make three contributions. First, coordinate-keyed structured retrieval improves land-use QA by 12.06--17.19 points on all 28{,}087 functional fMoW validation images. Second, retrieval's benefit and risk separate by verifiability: it reduces unsupported function and name claims, whereas fabricated visible-feature records cause 13.50--51.00-point losses; none of six arbitration prompts recovers the image-only baseline. Third, GeoArbiter preserves 84.69--87.15\% of the retrieval gain, cuts claim-level hallucination by 9.58--26.34\% under a source-blinded judge (19.28--45.19\% with the standard judge), and outperforms every prompt-level alternative. Historical OSM analysis further shows that apparent map--image age gaps mostly reflect mapping completion, not physical change.

\section{Related Work}
\label{sec:related}

\paragraph{Hallucination in general and RS MLLMs.}
Object hallucination in general-domain MLLMs is well documented \citep{pope}, and post-hoc systems such as Woodpecker verify generated claims after decoding \citep{woodpecker}. RS-specific studies broaden the taxonomy beyond object presence: RSHallu characterizes domain-specific knowledge errors \citep{rshallu}, while RADAR separates factual and logical failures and steers attention during decoding \citep{radar}. These methods improve how a model uses its existing visual and parametric evidence, but cannot supply a facility name or institutional function that the model never learned. Our setting targets precisely this missing-knowledge regime, while also measuring the new failure mode created when external knowledge contradicts perception.

\paragraph{Retrieval augmentation for remote sensing.}
RS-RAG retrieves encyclopedia entries through cross-modal embedding similarity and improves captioning and VQA for recognizable landmarks \citep{rsrag}. Other systems retrieve exemplars for SAR imagery \citep{sarrag}, image-derived context for aerial scenes \citep{aerorag}, or compose Earth-observation tools at inference time \citep{openearthagent}; CrisiSense-RAG combines reports and post-event imagery across evidence streams \citep{crisisense}. These approaches establish the value of inference-time context, but visual-similarity retrieval favors distinctive, documented sites and may return the wrong instance of an ordinary-looking facility. GeoArbiter instead uses coordinates as an exact retrieval key and evaluates ordinary function-defined locations at full fMoW scale. More importantly, we treat retrieval as a trust problem: the retrieved block is fallible evidence, not an oracle.

\paragraph{OSM in vision--language learning.}
OSM has been distilled into RS captions \citep{skyscript}, instruction data and multimodal alignment corpora \citep{lhrsbot,geolink}, map-based pre-training data \citep{fmowmm}, and rendered-tile domain adaptation \citep{osmda}. All primarily consume OSM before deployment. SkyScript is especially related because it classifies tags by whether overhead imagery can visually ground them, retaining groundable tags to improve caption fidelity \citep{skyscript}. We use the complementary inference-time selection: image-verifiable records are the dangerous ones when incorrect, so GeoArbiter withholds them and injects facts the image cannot adjudicate. Text-only coordinate-grounded systems \citep{geollm,spatialrag,georag} do not observe an image and therefore do not face this source conflict.

\paragraph{Knowledge conflict and adaptive retrieval.}
Multimodal conflict benchmarks construct contradictions among context, images, and parametric memory \citep{mmkcbench,segsub}; mechanistic studies further identify modality-specific preference patterns \citep{seeingoverrides,visiondefault}. Focus-on-vision prompting is often insufficient \citep{insightoversight}, and injected text can override visual evidence \citep{promptsoverride}. In text RAG, incorrect or outdated passages harm generation \citep{hoh}, motivating retrieval-quality estimation, query-complexity routing, reflection, and answer-attribution filtering \citep{selfrag,crag,adaptiverag,granurag}. Those methods ask whether retrieval is needed, relevant, or reliable. Verifiability is a different axis: an OSM record may be relevant and usually correct, yet still be the wrong evidence to expose when the image can directly settle the disputed attribute. To our knowledge, prior work has not used this property to select multimodal context or directly compared content-level with instruction-level arbitration.

\section{GeoArbiter}
\label{sec:method}

\subsection{Task formulation}
Given an RS image $I$, coordinates $(\phi,\lambda)$, footprint radius $r$, and instruction $q$, a frozen MLLM predicts $\hat{y}=\arg\max_y p_\theta(y\mid I,q,K)$. The knowledge block $K=\tau(\mathcal{S})$ deterministically textualizes retrieved records $\mathcal{S}$. Because no parameters are updated, the central design choice is which records to expose to the model.

\subsection{Coordinate-keyed retrieval}
\label{sec:retrieval}
Let $c(f)$ be the centroid of OSM feature $f$ and $\kappa(f)$ its primary tag key. We retrieve every feature within $\bar r=\mathrm{clamp}(r,300,1500)$\,m of $(\phi,\lambda)$, forming $\mathcal{R}$; full injection uses $\mathcal{S}=\mathcal{R}$. Dated Geofabrik extracts and a spatial grid index make retrieval reproducible without live API calls. On 461 images available through both routes, local and API feature counts correlate at $0.97$. Ablations add the nearest GeoNames toponym and dominant ESA WorldCover class \citep{worldcover}. Unlike embedding retrieval, coordinates identify ordinary as well as landmark locations and cannot return a visually similar but geographically unrelated site.

\subsection{Cross-modal verifiability and stratified injection}
\label{sec:stratmethod}
The pivotal property of a fact is whether the image could in principle confirm or refute it. We assign each OSM key a binary label $v$: keys expressing function or identity (\texttt{amenity}, \texttt{shop}, \texttt{tourism}, \texttt{military}, \texttt{office}) take $v=0$, whereas keys expressing physical structure (\texttt{building}, \texttt{highway}, \texttt{aeroway}, \texttt{natural}, \texttt{landuse}) take $v=1$; Appendix~\ref{app:keys} gives the complete map and textualization. Verifiability belongs to the attribute, not the object: a building may be visible while its institutional role is not.

For attribute class $a$, let $G(a)$ be the accuracy change from injecting $K$ instead of no knowledge. We test two predictions:
\begin{equation}
\begin{aligned}
\mathbb{E}[G\mid v{=}0] &> \mathbb{E}[G\mid v{=}1],\\
G(a\mid v{=}1,\mathrm{conflict}) &< 0 .
\end{aligned}
\label{eq:hyp}
\end{equation}
Thus, correct records may still provide a useful prior for visible attributes, but their marginal value should be smaller; when such a record conflicts with the image, its effect should be negative. A policy conditioned on $v$ should therefore outperform a fixed preference for either source.

Verifiability is distinct from both relevance and source reliability. A nearby runway record is relevant to a runway question, and OSM may be highly accurate on average, but a false runway record should not override an image in which the structure is visibly absent. Conversely, a school-function record cannot be confirmed from roof geometry even when the building itself is clear. We define $v$ by whether an attribute is in principle recoverable at the image's modality and scale, not by whether a particular model happens to answer it correctly. This makes the policy model-agnostic and prevents weak visual competence from being mistaken for inherent unverifiability.

\textbf{Stratified injection} enforces this policy on content, replacing $\mathcal{S}$ with the image-unverifiable subset
\begin{equation}
\mathcal{S}^{-}=\bigl\{f\in\mathcal{R}\;:\;v(\kappa(f))=0\bigr\},
\label{eq:filter}
\end{equation}
which retains 11.1\% of features. Records are textualized into a salience-ordered, 1{,}200-character block (named features, then tag counts). The filter requires no detector, extra model call, or per-image classifier.

The key-level map is intentionally conservative and auditable. It preserves OSM keys whose values primarily encode use, ownership, service, or identity and removes keys whose values primarily encode visible geometry, surface, or transport structure. It does not attempt to judge individual record correctness. This separation lets us test whether controlling exposure alone can outperform asking the same frozen model to reason about source trust after seeing every record.

\begin{center}
\footnotesize
\textbf{Policy summary.}\par\smallskip
\setlength{\tabcolsep}{3pt}
\begin{tabular*}{\columnwidth}{@{\extracolsep{\fill}}p{0.14\columnwidth}p{0.61\columnwidth}l@{}}
\toprule
Class & Representative OSM keys & Policy \\
\midrule
$v=0$ & \texttt{amenity}, \texttt{shop}, \texttt{tourism}, \texttt{military}, \texttt{office}, \texttt{name} & inject \\
$v=1$ & \texttt{building}, \texttt{highway}, \texttt{aeroway}, \texttt{natural}, \texttt{landuse} & withhold \\
\bottomrule
\end{tabular*}
\par\smallskip
\parbox{0.94\columnwidth}{\raggedright Function and identity are retained; visible structure is withheld. Boundary cases are analyzed in \S\ref{sec:stratified}.}
\end{center}

\subsection{Instruction-level arbitration (contrast arm)}
\label{sec:arbmethod}
The natural alternative leaves all records in the prompt and states the policy in language. We test \emph{image-first} (``if records conflict with what you see, trust the image''), \emph{database-first}, \emph{stratified} (image for visible structure; records for functions and names), a question-keyed rewrite, and two semantic-preserving paraphrases of the stratified instruction (Appendix~\ref{app:arb}). If frozen MLLMs can apply attribute-conditional trust, at least one of these six variants should reproduce Eq.~\ref{eq:filter}; none does (\S\ref{sec:probe}).

\subsection{Implementation}
Retrieval, textualization, and caching run on CPU. Warm-cache assembly takes 4.9\,ms per image (p95 12.6\,ms); injection adds 413 prompt tokens and 0.56\,s mean generation latency on Qwen2.5-VL-7B.

\begin{table*}[t]
\setlength{\abovecaptionskip}{4pt}
\setlength{\tabcolsep}{3.5pt}
\centering\small
\begin{tabular*}{\textwidth}{@{\extracolsep{\fill}}>{\raggedright\arraybackslash}p{0.12\textwidth}*{3}{>{\centering\arraybackslash}p{0.072\textwidth}}>{\centering\arraybackslash}p{0.02\textwidth}*{3}{>{\centering\arraybackslash}p{0.072\textwidth}}>{\centering\arraybackslash}p{0.02\textwidth}*{3}{>{\centering\arraybackslash}p{0.072\textwidth}}@{}}
\toprule
& \multicolumn{3}{c}{Land-use QA acc. (\%) $\uparrow$} & & \multicolumn{3}{c}{Hallucination rate (\%) $\downarrow$} & & \multicolumn{3}{c}{Conflict-cell acc. (\%) $\uparrow$}\\
\cmidrule(lr){2-4}\cmidrule(lr){6-8}\cmidrule(lr){10-12}
Condition & Qwen & InternVL & LLaVA & & Qwen & InternVL & LLaVA & & Qwen & InternVL & LLaVA\\
\midrule
none          &62.78 &58.97 &70.53 & &4.89 &4.22 &6.07 & &88.75 &75.75 &87.75\\
coords        &64.07 &60.63 &71.27 & & --- & --- & --- & & --- & --- & ---\\
rag           & \best{79.30} & \best{76.16} & \best{82.59} & & \best{2.36} & \best{3.21} & \best{4.64} & &72.00 &62.25 &36.75\\
rag-filtered  &76.77 &73.55 &81.04 & &2.68 &3.30 &4.90 & & \best{93.25} & \best{90.50} & \best{91.25}\\
\bottomrule
\end{tabular*}
\caption{\textbf{Core results across three open MLLMs.} Left: land-use QA accuracy on the full fMoW functional validation split ($n{=}28{,}087$ per cell). Middle: claim-level hallucination rate (\textsc{contradicted} share) on 3{,}000 scene descriptions ($\approx$8.5--9 claims each). Right: existence-probe conflict-cell accuracy, where a \emph{fabricated} record is injected and ground truth is \emph{no} ($n{=}1{,}069$; full per-cell probe in Table~\ref{tab:probe_full}). Full injection (\emph{rag}) is nominally best on the clean tasks but \emph{collapses} once a record is wrong (Qwen $79.30$ QA yet $72.00$ conflict; LLaVA down to $36.75$); stratified injection (\emph{rag-filtered}) keeps $85\%$ of the QA gain and, by withholding image-verifiable records, is the only condition robust in conflict, exceeding even the no-knowledge cell. \emph{coords} (image plus raw latitude/longitude) is a QA-only baseline. All rag$-$none and filtered$-$rag differences are significant ($p{<}0.001$ paired bootstrap for QA; $p{<}10^{-4}$ for hallucination and the conflict cell). Per-model and per-type breakdowns are in Appendix~\ref{app:results}.}
\label{tab:main}\label{tab:halluc}
\end{table*}

\section{Experimental setup}
\label{sec:setup}

\paragraph{Data.}
Our main corpus is the complete fMoW functional validation split \citep{fmow}: 28{,}087 images in 32 function-defined categories, with WGS84 coordinates and 2002--2017 acquisition times. We avoid OSM-derived QA and captions because evaluating OSM-grounded generation against OSM-derived labels would be circular. fMoW and our CORINE-derived external labels are independent of the injected knowledge.

\paragraph{Models.}
Three open MLLMs spanning architecture families: Qwen2.5-VL-7B-Instruct \citep{qwen25vl}, InternVL3-8B \citep{internvl3}, and LLaVA-OneVision-7B \citep{llavaonevision}, all frozen, greedy decoding.

\paragraph{Conditions.}
\emph{none} (image only); \emph{coords} (plus raw coordinates); \emph{rag} (full OSM, $\mathcal{S}=\mathcal{R}$); and \emph{rag-filtered} (GeoArbiter, $\mathcal{S}=\mathcal{S}^{-}$). Additional comparisons include the six arbitration instructions, Wikipedia geosearch, GeoNames, WorldCover, all sources, and a reproduction of RS-RAG retrieval.

\paragraph{Land-use QA and description.}
Land-use QA is four-way fMoW classification. Distractors are sampled with a fixed per-image seed, so all models and conditions receive identical options; we report accuracy with paired bootstrap tests (10k resamples). For open generation, models produce 3--5 sentence descriptions on a seeded 3{,}000-image subset. A frozen Qwen2.5-7B-Instruct judge decomposes each output into atomic claims (8.5--9 per description on average), assigns a type (\textsc{function}, \textsc{name}, physical \textsc{context}, or other), and labels it \textsc{supported}, \textsc{contradicted}, or \textsc{unverifiable} against the image label and retrieved reference. Hallucination rate is the \textsc{contradicted} share; prompts and the judge template are in Appendices~\ref{app:prompts} and~\ref{app:judge}, and human validation is in \S\ref{sec:validity}.

\paragraph{Verifiable-existence probe.}
To isolate the regime in which the image should dominate, we build 1{,}069 yes/no items from 400 images. The \emph{conflict} cell adds a fabricated record for a manually verified absent, visually distinctive feature. \emph{Control-absent} asks about the same absence without fabrication, while \emph{control-present} asks about a manually verified present feature. The former separates susceptibility to injected records from ordinary false positives; the latter reveals instructions that improve conflict accuracy merely by shifting the response prior toward \emph{no}. Historical and current OSM snapshots ensure that each selected feature's map status is persistent, but ground truth comes from image inspection rather than OSM absence.

\paragraph{Derived metrics.}
The \emph{retained gain} $\rho=(a_{\mathrm{filt}}-a_{\varnothing})/(a_{\mathrm{rag}}-a_{\varnothing})$ measures how much full-retrieval QA gain survives filtering. Because instructions can shift the yes/no prior, we also report probe balanced accuracy $\mathrm{BA}=\tfrac12(a_{\mathrm{conf}}+a_{\mathrm{pres}})$ rather than interpreting the conflict cell alone.

\paragraph{External benchmark.}
We construct 2{,}000 four-way land-cover questions from a pan-European BigEarthNet subset \citep{bigearthnet} via GEO-Bench \citep{geobench}, preserving patch coordinates. RSHBench \citep{radar} and RSHalluEval \citep{rshallu} were unreleased at submission time and are used only as qualitative context.

\section{Results}
\label{sec:results}

\subsection{Structured retrieval lifts land-use QA, and content is what matters}
\label{sec:mainqa}

Coordinate-keyed OSM injection raised accuracy by $+16.52$, $+17.19$, and $+12.06$ points for Qwen2.5-VL, InternVL3, and LLaVA-OneVision (Table~\ref{tab:main}; all $p{<}0.001$). Raw coordinates added at most $+1.66$ points; retrieval's gain was 10.36--16.30$\times$ larger, ruling out location disclosure. Three controls further localize the gain: an image-blind OSM-tag decoder tops out at 70.6\%, and blanking the image or pairing it with a \emph{different} image's knowledge drops accuracy back toward \emph{none}, so the gain needs coordinate-matched knowledge bound to the image, not tag leakage (Appendix~\ref{app:modality}); at a matched record budget, verifiability-selected injection also retains far more of it than random, frequency, or inverted filtering (Appendix~\ref{app:controls}). LLaVA-OneVision, the strongest baseline, gained least, as expected if retrieval fills a knowledge gap rather than adding a generic prior.

\subsection{Injection reduces claim-level hallucination where verifiability predicts}
\label{sec:halluc}

\begin{figure}[t]
\centering
\includegraphics[width=\columnwidth]{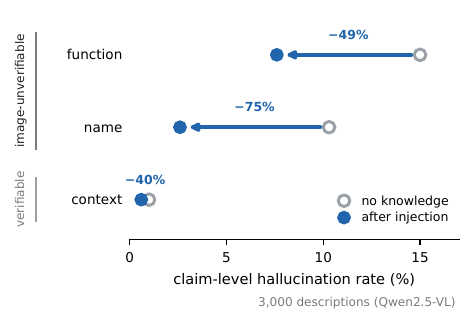}
\caption{\textbf{Injection helps most on image-unverifiable claims.} Claim-level hallucination rates over 3{,}000 Qwen2.5-VL descriptions before retrieval (hollow) and after full OSM injection (filled). Arrows report relative reductions: function and name claims show the largest absolute gains, while visually verifiable context starts near the floor.}
\label{fig:asym}
\end{figure}

Full injection cut claim-level hallucination by 51.74\%, 23.93\%, and 23.56\% (Table~\ref{tab:main}), with stable claim counts. Reductions concentrated on image-unverifiable content (Figure~\ref{fig:asym}; all models: Table~\ref{tab:htype_full}). For Qwen2.5-VL, \textsc{function} fell from 15.01\% to 7.63\%, and \textsc{name} from $10.29\%$ to $2.59\%$ despite 2--5$\times$ more naming claims; physical-\textsc{context} began at 1.01\% and changed little. The \textsc{unverifiable} share also fell from 51.76\% to 32.71\%, indicating more checkable rather than merely shorter descriptions. These rates use a judge that sees the injected records; under a source-blinded judge the \emph{none}$\to$filtered reduction is 9.58--26.34\% (vs.\ 19.28--45.19\%) and absolute rates roughly double, so we treat injected-reference values as a lower bound and report the blinded figure (Appendix~\ref{app:blind}).

\subsection{Retrieval changes many answers, and the database usually wins on merit}
\label{sec:conflicts}
Injection changed 27\%, 28\%, and 18\% of model answers. An answer change is a superset of a genuine image--database conflict---it also captures cases where the database merely supplies information the image cannot show---so we read it as a retrieval-induced answer change rather than a verified natural conflict. Among these changes, records corrected the image-only answer 72--77\% of the time and misled it 10--11\%. The most frequently rescued classes---places of worship, emergency services, gas stations, and schools---are defined by functions that are invisible from orbit (Figure~\ref{fig:cases}a). Natural function conflicts therefore favor the database; the controlled existence probe tests the complementary, visually verifiable regime.

\subsection{The verifiable side: fabricated records fool every model, and instructions cannot fix it}
\label{sec:probe}

\begin{table*}[t]\centering\small
\begin{tabular*}{\textwidth}{@{\extracolsep{\fill}}lccccccccc@{}}
\toprule
acc. (\%)\,$\uparrow$ & \multicolumn{3}{c}{Qwen2.5-VL} & \multicolumn{3}{c}{InternVL3} & \multicolumn{3}{c}{LLaVA-OV}\\
\cmidrule(lr){2-4}\cmidrule(lr){5-7}\cmidrule(lr){8-10}
Condition & conf & abs & pres & conf & abs & pres & conf & abs & pres\\
\midrule
no knowledge &88.75 &87.66 &61.40 &75.75 &76.83 &70.22 &87.75 &89.67 &68.38\\
rag (fabricated) &72.00 &96.22 &61.76 &62.25 &92.44 &70.59 &36.75 &94.46 &75.00\\
\quad + image-first &76.25 &96.73 &62.50 &70.00 &93.45 &65.81 &42.50 &94.96 &73.53\\
\quad + database-first &81.00 &96.98 &57.35 &71.25 &93.95 &62.13 &41.00 &95.21 &73.16\\
\quad + stratified &75.50 &96.98 &62.87 &75.00 &94.96 &62.50 &43.50 &95.21 &72.06\\
\quad + question-keyed &73.00 &95.97 &65.07 &70.00 &94.46 &64.71 &43.25 &94.71 &73.53\\
stratified injection & \best{93.25} &94.71 &66.91 & \best{90.50} &91.44 &70.59 & \best{91.25} &92.95 &72.79\\
\bottomrule
\end{tabular*}
\caption{Existence probe, per-cell accuracy in percent (conf: conflict, GT \emph{no}; abs: control-absent, GT \emph{no}; pres: control-present, GT \emph{yes}; $n{=}1{,}069$). Balanced accuracy is $\tfrac12$(conf${+}$pres). Fabricated records cost the conflict cell; no instruction restores it, and stratified injection exceeds the no-knowledge conflict cell for every model.}
\label{tab:probe_full}
\end{table*}

Fabricated records cut conflict-cell accuracy from 88.75\% to 72.00\% for Qwen2.5-VL, 75.75\% to 62.25\% for InternVL3, and 87.75\% to 36.75\% for LLaVA-OneVision (all $p{<}10^{-4}$): losses of 13.50--51.00 points. This is the multimodal counterpart of context over-reliance in text RAG \citep{hoh,promptsoverride}.

Arbitration prompts partially mitigated conflict but had two systematic defects. First, \emph{leakage}: every image-trust instruction reduced all-function QA by 0.50--3.25 points on Qwen2.5-VL; the bootstrap probability of a decrease exceeded 0.999 for the worst variant. Only database-first helped ($+1.50$, $p{=}0.049$). Second, \emph{response bias}: instructions shifted the yes/no prior toward \emph{no}, improving conflict scores while reducing control-present accuracy. Consequently, no instruction recovered the no-knowledge balanced accuracy (Table~\ref{tab:probe_full}; full grid: Table~\ref{tab:instr_full}). Frozen models do not reliably execute attribute-conditional trust from language alone.

\subsection{Stratified injection wins both sides}
\label{sec:stratified}
\begin{figure*}[t]
\centering
\includegraphics[width=0.82\textwidth]{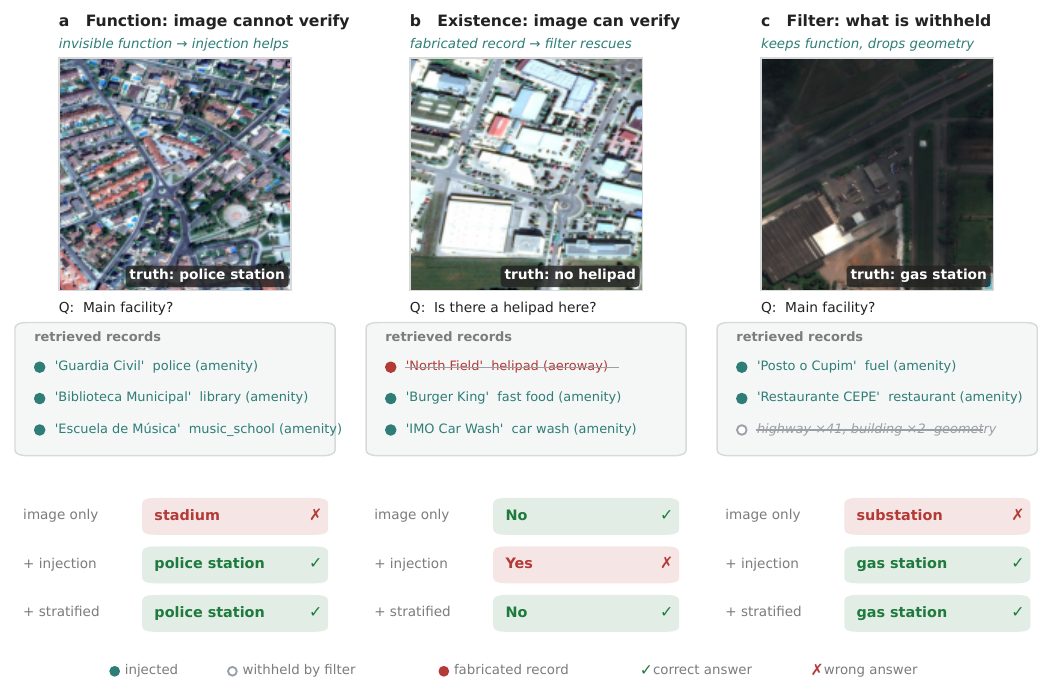}
\caption{\textbf{Three cases} (unedited Qwen2.5-VL-7B outputs).
\textbf{a}, Function is invisible from orbit: the model guesses \emph{stadium};
a retrieved \emph{police} record (`Guardia Civil') settles it, and stratified
injection keeps that record. \textbf{b}, Existence is visible: a planted
\emph{helipad} record flips the answer to \emph{yes}, and withholding the
image-verifiable record restores \emph{no}. \textbf{c}, The filter drops the
bulk road and building geometry while injecting the two function records that
identify the site.}
\label{fig:cases}
\end{figure*}

Content-level filtering discarded 88.9\% of features yet retained 84.69--87.15\% of QA gain and 81.82--91.09\% of hallucination reduction, with at most a 0.32-point cost (Table~\ref{tab:main}). Probe conflict accuracy reached 93.25\%, 90.50\%, and 91.25\%; balanced accuracy was 80.08\%, 80.54\%, and 82.02\%, above the no-knowledge baseline for every model (Table~\ref{tab:probe_full}). Figure~\ref{fig:cases}b--c shows GeoArbiter withholding fabricated and structural records while retaining function records. The diagnostic exception is visible \texttt{amenity=parking}, which survives the filter and fools models in 83--92\% of cases. Robustness comes from withholding records, not model-side arbitration: at a matched record budget, keeping the image-\emph{verifiable} records instead (inverted filter) drops conflict accuracy to 20.5--52.8\%, below full injection, whereas random selection stays robust---the direction of selection, not the reduced context, is the lever (Appendix~\ref{app:controls}).

\subsection{Measurement validity: human agreement}
\label{sec:validity}
Two annotators assessed a stratified 198-claim sample (balanced across judge labels, models, and conditions), blind to the judge. Initial agreement was 78.3\% ($\kappa{=}0.39$); most of the 43 disagreements concerned two underspecified cases (hedged wrong guesses; claims restating a reference category). After clarifying that hedging does not excuse asserted content and that a correct restatement is not a contradiction, the annotators resolved 28 cases and excluded 15 perceptual ones. Against the 183 adjudicated labels the judge reached 84.2\% agreement ($\kappa{=}0.61$, recall 0.88, precision 0.60), so we treat absolute rates as conservative upper bounds and emphasize paired differences. The modest initial agreement remains a limitation (\S\ref{sec:limitations}).

\subsection{Baselines and source ablations}
\label{sec:baselines}

On a seeded 3{,}000-image subset (full table in Appendix~\ref{app:results}, Table~\ref{tab:ablation_full}), structured OSM (ours) reaches 80.57/76.50/83.33\% vs.\ 69.13/65.40/73.43\% for Wikipedia GeoSearch \citep{wikigeosearch}, 65.83/62.77/72.53\% for GeoNames \citep{geonames}, and 58.50/60.07/70.43\% for a faithful RS-RAG reproduction \citep{rsrag} over its released 14{,}820-landmark base \citep{openstreetmap}. RS-RAG was neutral to harmful on ordinary scenes ($-5.30$ points for Qwen): retrieved landmarks looked similar ($0.90{\pm}0.03$) but had the wrong identity. Wikipedia GeoSearch added 2.96--7.17 points; OSM gains were 2.55--4.34$\times$ larger. GeoNames added 2.03--4.54 points and WorldCover \citep{worldcover} 1.43--5.30, while combining them with OSM changed accuracy by only $-0.63$ to $+0.76$ points.

\subsection{External validity and cost}
On BigEarthNet land-cover QA, injection improved all three models by 7.95, 4.25, and 4.10 points (Table~\ref{tab:ben_full}). These smaller gains are consistent with Eq.~\ref{eq:hyp}: land cover is largely image-verifiable, leaving less missing knowledge to supply. Deployment overhead is 4.9\,ms warm-cache assembly, 413 prompt tokens, and 0.56\,s mean generation latency on Qwen2.5-VL-7B.

\section{Analysis: why the axis is verifiability, not time}
\label{sec:analysis}

\subsection{Why content-level arbitration succeeds}
\label{sec:whycontent}
The two task families expose an incompatibility no global source preference can solve. Function-defined fMoW questions favor the records because the decisive attribute is usually invisible, so image-first language removes useful evidence and lowers accuracy; existence questions under conflict favor the image because the disputed structure is directly observable, so database-first behavior produces false positives. A correct policy must switch at the attribute level, not the image, question, or database level.

The instruction experiments show why stating this switch is insufficient: even the stratified and question-keyed prompts leave the conflicting record in context and require the frozen model to identify the disputed attribute, classify its verifiability, and inhibit a salient textual assertion. Their leakage on function QA shows the instruction acts partly as a global image-trust prior; the drop on control-present items shows an added \emph{no}-response bias. Balanced accuracy exposes both, which reporting the conflict cell alone would obscure.

GeoArbiter instead compiles the policy into the input: an image-verifiable attribute's record is absent and cannot compete with perception, while an unverifiable one remains available without a weakening instruction. This explains the otherwise striking combination---retaining only 11.1\% of records preserves 84.69--87.15\% of the land-use gain, while conflict balanced accuracy exceeds the image-only baseline for every model. The filter does not make the MLLM a better arbiter; it removes the need for arbitration.

\subsection{What the controlled conflict establishes}
The fabricated-record probe is not meant to estimate the prevalence of OSM errors; it is a controlled intervention on the evidence channel, holding image and question fixed while the asserted record is added or removed. The control-absent cell estimates the model's ordinary false-positive tendency, and control-present measures whether an intervention shifts the response prior. The large conflict-only loss under full injection thus identifies exposure to the contradictory record as the mechanism, not image difficulty or a general \emph{yes} tendency. Retrieval-induced answer changes give the complementary ecological result---when disagreements concern function, records usually help---so the two analyses locate the boundary of useful grounding rather than merely showing retrieval can help or hurt.

\subsection{Temporal mismatch does not explain the gap}
Archival imagery (2002--2017) paired with a current map suggests temporal staleness as a rival explanation. Reconstructing acquisition-time OSM, we compute the map--image divergence $\Delta=1-(|F_t\cap F_0|-|C|/2)/|F_t\cup F_0|$ ($F_t$/$F_0$: historical/current features; $C$: shared features with changed tags). Retrieval gain was flat across $\Delta$ quantiles for both pilot models ($-0.01$ to $-0.06$, $p{\geq}0.61$), and current maps outperformed sparser time-matched snapshots; a blinded audit explains why: 93.9\% of features mapped after acquisition were already visible in the older image, and only about 1 in 22 scenes showed new construction. On this corpus the OSM--image delta therefore primarily measures mapping completion, not physical change; the audit is modest and we do not claim this generalizes beyond it.

\subsection{Where the filter still costs}
Key-level stratification drops useful evidence expressed through otherwise verifiable keys, such as hangars under \texttt{aeroway} and offices under \texttt{building}, accounting for the 12.85--15.31\% loss in retained gain; categories with strong \texttt{amenity} evidence lose nothing. The natural refinement is an attribute-level policy over tag values or question--attribute pairs---recovering functional evidence under \texttt{building}/\texttt{aeroway} while withholding visibly checkable values such as \texttt{amenity=parking}---still within a deterministic, training-free schema map.

\section{Conclusion}
\label{sec:conclusion}
Coordinate-keyed geographic knowledge improves RS MLLMs but harms when it contradicts image-verifiable attributes. Cross-modal verifiability predicts this boundary, which frozen models cannot enforce from instructions; GeoArbiter instead filters content, keeping 84.69--87.15\% of the QA gain and 81.82--91.09\% of the hallucination reduction while staying robust to fabricated records. Verifiability-guided selection is a strong, training-free baseline for grounding VLMs in fallible structured knowledge.

\section*{Limitations}
\label{sec:limitations}
Our verifiability map is a key-level approximation: a parking lot is an \texttt{amenity} but plainly visible, while some useful functional evidence appears under structural keys. The automatic hallucination judge has only moderate agreement with adjudicated human labels ($\kappa{=}0.61$), and initial human agreement was also modest ($\kappa{=}0.39$); absolute rates should therefore be read as conservative upper bounds, although all comparisons use the same judge. Probe conflicts are constructed rather than naturally observed, so they measure susceptibility under controlled contradictions rather than prevalence in deployed OSM. Results cover fMoW, BigEarthNet, English prompts, and three 7--8B open models; RS-tuned or larger models may follow different source priors. Two unreleased RS hallucination benchmarks limit comparison.

\section*{Ethical Considerations}
This work uses public datasets, open-weight models, and geographic records under their respective licenses. Imagery comes from fMoW and BigEarthNet under their released research terms, and we use SkyScript images only (not its OSM-derived captions). Structured knowledge comes from OpenStreetMap (ODbL~1.0), GeoNames (CC-BY~4.0), and ESA WorldCover (CC-BY~4.0). Because ODbL is an attribution and share-alike licence, any OSM-derived caches we release carry the OpenStreetMap attribution and the same licence, and we redistribute derived features rather than raw third-party imagery. Retrieval operates only on coordinates already distributed with the benchmarks and describes public infrastructure rather than individuals. Nevertheless, joining overhead imagery to precise geographic records can support surveillance as well as civilian land-use analysis; deployments should therefore apply access controls and purpose-specific review.

Two project members performed the claim and existence checks. Because the task labels model outputs rather than studying the annotators, it is not human-subjects research and required no ethics approval; the annotators participated voluntarily, with informed consent, and were uncompensated. They viewed no sensitive content. The annotation items and guidelines accompany the submission; generated descriptions were used only for verification and are not redistributed as ground truth.

\bibliography{custom}

\clearpage
\appendix
\raggedbottom
\numberwithin{table}{section}
\makeatletter
\setlength{\@dblfptop}{0pt}
\makeatother

\section{Reproducibility details}
\label{app:repro}
All prompts (task instructions, the six arbitration wordings of Appendix~\ref{app:arb}, judge template), the knowledge textualization format, per-image caches, and Slurm run scripts are released. Local OSM extraction uses dated Geofabrik continental extracts (2026-07-25) with centroid-in-bbox semantics matching the ohsome centroid endpoint used in the pilot; validation against 461 API-fetched images gives count correlation 0.97 (median local/API ratio 0.92; relations are skipped). The RS-RAG reproduction and its documented deviations (unreleased monthly-knowledge and image-description chunks) are detailed in the released notes. Table~\ref{tab:repro} lists the remaining settings needed to reproduce every number.

\begin{table}[H]
\centering\footnotesize
\setlength{\tabcolsep}{4pt}
\begin{tabular*}{\columnwidth}{@{\extracolsep{\fill}}p{0.27\columnwidth}p{0.66\columnwidth}@{}}
\toprule
Component & Setting\\
\midrule
Backbones (HF rev.) & Qwen2.5-VL-7B-Instruct (\texttt{cc59489}), InternVL3-8B (\texttt{853e3a7}), llava-onevision-qwen2-7b-ov-hf (\texttt{0d50680})\\
Judge & Qwen2.5-7B-Instruct (\texttt{a09a354})\\
Decoding & greedy (\texttt{do\_sample=False}); \texttt{max\_new\_tokens} 256 (judge 512); single pass\\
Image preproc. & Qwen $256$--$1280{\times}28^2$ px; InternVL $448$px dynamic tiling ($\leq6$); LLaVA default processor; all RGB\\
Answer parsing & first \texttt{A}--\texttt{D} letter in the response; unparsable counts wrong\\
Retrieval & centroid within $\bar r{=}\mathrm{clamp}(r,300,1500)$\,m; knowledge block $\leq1{,}200$ chars (App.~\ref{app:keys})\\
OSM snapshot & Geofabrik 2026-07-25; ohsome acquisition-time snapshots for the temporal probe\\
CIs & location-clustered ($\texttt{category\_LOC}$) block bootstrap, $10{,}000$ resamples, seed 0; QA paired by image\\
Seeds & QA options seeded per image id; existence-probe construction seed 42\\
Released & code, configs, per-image caches, judged outputs, Slurm scripts\\
\bottomrule
\end{tabular*}
\caption{Reproducibility settings (companion to the released code and caches).}
\label{tab:repro}
\end{table}

\section{Arbitration instruction wordings}
\label{app:arb}
The instruction-level arbitration arm (\S\ref{sec:arbmethod}) appends one of six trust instructions to the knowledge block, listed verbatim below. Instructions~1--2 name a global winner; instruction~3 states the verifiability rule keyed on object class, and instructions~4--5 are two paraphrases of it that control for wording; instruction~6 (question-keyed) conditions on the disputed attribute rather than the object class. The results in \S\ref{sec:probe} hold across all six.

\begin{enumerate}
\item \textbf{Image-first.} ``If the database records conflict with what you see in the image, trust the image.''
\item \textbf{Database-first.} ``If the database records conflict with what you see in the image, trust the database records.''
\item \textbf{Stratified (object-keyed).} ``If the database records conflict with what you see, decide by the type of information: for physically visible things (buildings, roads, water, runways), trust the image; for functions and names that cannot be verified visually (e.g.\ whether a building is a school or an office, place names), trust the database records.''
\item \textbf{Stratified, paraphrase 1.} ``When sources disagree, apply this rule: believe the image for physically observable features such as buildings, roads, water bodies and runways; believe the database for attributes the image cannot verify, such as a building's function or a place name.''
\item \textbf{Stratified, paraphrase 2.} ``Resolve conflicts by information type: visual evidence wins for anything directly observable (structures, roads, water, runways); database records win for visually unverifiable facts (facility functions, names).''
\item \textbf{Question-keyed.} ``If the database records conflict with what you see, decide by what is in dispute: if the question is whether something exists or how it is laid out (a runway, a pool, a building being there), trust the image; if the question is what a facility is for or what it is called (school vs office, place names), trust the database records.''
\end{enumerate}

\section{Prompt templates}
\label{app:prompts}
Braced placeholders are filled per image; the six arbitration wordings are in Appendix~\ref{app:arb}.

\paragraph{Land-use QA.}
``Look at the satellite image and answer the question. Question: What is the primary land use or function of the main facility shown in this image? Options: \{options\} Answer with the letter of the single best option.''

\paragraph{Scene description.}
``Describe this satellite image in 3--5 sentences: the main facility or land use, notable objects, and spatial layout. Only state what you can support; if the location or a name is uncertain, say so.''

\paragraph{Knowledge preamble} (prepended for \emph{rag}).
``Geographic database records for this image's location (may be outdated or incomplete): \{knowledge\}''

\paragraph{Coordinate preamble} (for \emph{coords}).
``This image was taken at latitude \{lat\}, longitude \{lon\}.''

\section{Verifiability keys and textualization}
\label{app:keys}

\paragraph{Verifiability map.}
A retrieved feature is image-\emph{un}verifiable ($v{=}0$, injected) when its primary OSM key is one of \texttt{amenity}, \texttt{shop}, \texttt{tourism}, \texttt{military}, \texttt{office}, \texttt{healthcare}, \texttt{craft}, \texttt{religion}, \texttt{operator}, \texttt{brand}. All other keys (\texttt{building}, \texttt{highway}, \texttt{aeroway}, \texttt{natural}, \texttt{landuse}, \texttt{railway}, \texttt{man\_made}, \texttt{leisure}, \texttt{power}, \texttt{waterway}, \texttt{sport}, \dots) are image-verifiable ($v{=}1$) and withheld by stratified injection, which retains 11.1\% of retrieved features on our data.

\paragraph{Textualization.}
Records are serialized into a character-budgeted block ($\leq1{,}200$ chars) in salience order: an optional land-cover line (WorldCover) and nearest-toponym line (GeoNames), then named features, then per-type feature counts. Features are ordered by key: \texttt{aeroway}, \texttt{amenity}, \texttt{railway}, \texttt{military}, \texttt{man\_made}, \texttt{leisure}, \texttt{tourism}, \texttt{shop}, \texttt{power}, \texttt{waterway}, \texttt{natural}, \texttt{landuse}, \texttt{highway}, \texttt{building}, \texttt{sport}.

\section{Hallucination judging protocol}
\label{app:judge}
A frozen Qwen2.5-7B-Instruct judge decomposes each description into atomic claims and labels each with a type and verdict: ``You are auditing a satellite-image description for factual errors. Reference facts: the image shows a facility of category \{category\}; \{knowledge block\}. [description]. List every distinct factual claim; for each output one line \texttt{CLAIM: $\ldots$ | TYPE: FUNCTION/NAME/CONTEXT/OTHER | VERDICT: SUPPORTED/CONTRADICTED/UNVERIFIABLE}. Judge strictly against the reference facts; use \textsc{unverifiable} when the reference is silent.'' The hallucination rate is the \textsc{contradicted} share of all claims; human validation of the judge is in \S\ref{sec:validity}.

\section{Extended results}
\label{app:results}
All tables are recomputed from the released per-image outputs.
\setlength{\intextsep}{5pt}
\setlength{\abovecaptionskip}{4pt}

\begin{table}[H]
\centering\small
\begin{tabular*}{\columnwidth}{@{\extracolsep{\fill}}lrr@{}}
\toprule
Instruction & Qwen2.5-VL & InternVL3\\
\midrule
no instruction        &81.00          &76.25\\
image-first           &80.50 ($-0.50$) &74.50 ($-1.75$)\\
database-first        &82.50 ($+1.50$) &74.75 ($-1.50$)\\
stratified            &77.75 ($-3.25$) &75.25 ($-1.00$)\\
\quad paraphrase 1    &78.25 ($-2.75$) &75.25 ($-1.00$)\\
\quad paraphrase 2    &79.50 ($-1.50$) &75.50 ($-0.75$)\\
question-keyed        &79.25 ($-1.75$) &74.25 ($-2.00$)\\
\bottomrule
\end{tabular*}
\caption{Instruction-level arbitration accuracy (\%) on the all-function QA pilot ($n{=}400$); parenthesized differences are percentage-point changes from no-instruction rag. Every image-trust instruction hurts; only database-first helps. LLaVA-OV was not run in the pilot sweep.}
\label{tab:instr_full}
\end{table}

\begin{table}[H]
\centering\small
\begin{tabular*}{\columnwidth}{@{\extracolsep{\fill}}llrrr@{}}
\toprule
& & \multicolumn{3}{c}{Hallucination rate (\%) $\downarrow$}\\
\cmidrule(lr){3-5}
Model & & \textsc{func.} & \textsc{name} & \textsc{ctx.}\\
\midrule
\multirow{2}{*}{Qwen2.5-VL} & none &15.01 &10.29 &1.01\\
 & rag &7.63 &2.59 &0.57\\
\multirow{2}{*}{InternVL3} & none &14.55 &8.01 &0.73\\
 & rag &11.58 &6.75 &0.58\\
\multirow{2}{*}{LLaVA-OV} & none &18.36 &6.46 &1.40\\
 & rag &16.16 &5.00 &1.01\\
\bottomrule
\end{tabular*}
\caption{Claim-level hallucination rate by knowledge type (\textsc{contradicted} share of that type's claims). Injection helps most on the least verifiable types; Figure~\ref{fig:asym} plots Qwen.}
\label{tab:htype_full}
\end{table}

\begin{table}[H]
\centering\footnotesize
\setlength{\tabcolsep}{1.6pt}
\begin{tabular*}{\columnwidth}{@{\extracolsep{\fill}}lrrrrrrr@{}}
\toprule
& \multicolumn{7}{c}{Accuracy (\%) $\uparrow$}\\
\cmidrule(lr){2-8}
Model & none & RS-RAG & Wiki & WC & GN & OSM & all\\
\midrule
Qwen     &63.80 &58.50 &69.13 &65.23 &65.83 &80.57 &\best{81.33}\\
InternVL &58.23 &60.07 &65.40 &63.53 &62.77 &\best{76.50} &76.23\\
LLaVA    &70.47 &70.43 &73.43 &72.43 &72.53 &\best{83.33} &82.70\\
\bottomrule
\end{tabular*}
\caption{Retrieval baselines and single-source ablations (3{,}000-image subset; full version of the baselines in \S\ref{sec:baselines}). WC: WorldCover-only; GN: GeoNames-only; OSM: coordinate-keyed structured OSM (ours); all: OSM${+}$GN${+}$WC. Structured OSM dominates; adding GeoNames and WorldCover changes little.}
\label{tab:ablation_full}
\end{table}

\begin{table}[H]
\centering\small
\begin{tabular*}{\columnwidth}{@{\extracolsep{\fill}}lrrr@{}}
\toprule
Model & none & rag & $\Delta$ (points)\\
\midrule
Qwen2.5-VL &43.85 &51.80 &$+7.95$\\
InternVL3  &50.45 &54.70 &$+4.25$\\
LLaVA-OV   &47.50 &51.60 &$+4.10$\\
\bottomrule
\end{tabular*}
\caption{External BigEarthNet land-cover QA accuracy (\%; $n{=}2{,}000$). Gains are smaller than on fMoW, consistent with Eq.~\ref{eq:hyp}: land-cover classes are largely image-verifiable.}
\label{tab:ben_full}
\end{table}

\section{Verifiability vs.\ retrieval budget}
\label{app:controls}
To separate \emph{what} is injected from \emph{how much}, we compare the verifiability filter against three controls that keep the \emph{same} number of records per image (11.4\% of the retrieved set, matched to the record): \emph{random} keeps a random subset; \emph{freq} keeps the commonest tag-key records; \emph{inverted} keeps image-\emph{verifiable} records (the complement of our rule). Table~\ref{tab:controls} scores all of them. On QA accuracy the filter retains $\approx$85\% of the full-injection gain while the budget-matched controls retain only 28--43\%; on the conflict probe the filter and random stay robust but \emph{inverted collapses below full injection}, because at identical budget it keeps the fabricated image-verifiable record (fabrication survival: filter ${\sim}0/400$, random $83/400$, inverted $388/400$). Selection by verifiability, not context size, is the lever on both sides. A rule-based OSM-tag$\rightarrow$label decoder that never sees the image (generous keyword matching) tops out at 70.6\% on the same four-way QA, below every model's \emph{rag} accuracy, so the gain is not the tags spelling out the label. All confidence intervals in Tables~\ref{tab:main} and \ref{tab:controls} are location-clustered block bootstrap (resampling whole \texttt{category\_LOC} groups); every headline gap survives clustering at $p{<}10^{-4}$.

\begin{table}[H]
\centering\footnotesize
\setlength{\tabcolsep}{3pt}
\begin{tabular*}{\columnwidth}{@{\extracolsep{\fill}}lrrrcrrr@{}}
\toprule
& \multicolumn{3}{c}{QA acc (\%)} && \multicolumn{3}{c}{conflict (\%)}\\
\cmidrule(lr){2-4}\cmidrule(lr){6-8}
condition & Qw & In & Lv && Qw & In & Lv\\
\midrule
none              &62.8 &59.0 &70.5 && 88.8 &75.8 &87.8\\
full injection    &79.3 &76.2 &82.6 && 72.0 &62.3 &36.8\\
random (matched)  &69.9 &68.3 &76.1 && 90.5 &84.0 &83.8\\
freq (matched)    &67.4 &66.3 &74.4 && --   &--   &--\\
inverted (matched)&67.5 &66.4 &74.6 && 48.2 &52.8 &20.5\\
filtered (ours)   &76.8 &73.5 &81.0 && \best{93.2} &\best{90.5} &\best{91.2}\\
\bottomrule
\end{tabular*}
\caption{Budget-matched control filters (Qw: Qwen2.5-VL, In: InternVL3, Lv: LLaVA-OV). All non-\emph{none}/\emph{full} rows keep the same 11.4\% record budget. Conflict is the existence-probe conflict cell ($n{=}1{,}069$); \emph{freq} was not run on the probe. Full injection is best on clean QA but collapses in conflict; the verifiability filter is the only budget-respecting condition strong on both.}
\label{tab:controls}
\end{table}

\section{Modality isolation}
\label{app:modality}
Table~\ref{tab:modality} isolates the image's contribution. \emph{knowledge only} replaces the image with a blank frame; \emph{shuffle image} pairs each real image with a \emph{different} image's retrieved knowledge. Text alone carries signal (\emph{knowledge only} $>$ \emph{none}), but the real-image condition beats it, and \emph{shuffle image} collapses back to the no-knowledge baseline: the gain requires coordinate-matched knowledge bound to the image, not free-floating text, ruling out pure label-leakage from the tags. Permuting the coordinates likewise left accuracy at the \emph{none} level (InternVL3 $60.3$).

\begin{table}[H]
\centering\small
\setlength{\tabcolsep}{5pt}
\begin{tabular*}{\columnwidth}{@{\extracolsep{\fill}}lrrr@{}}
\toprule
& \multicolumn{3}{c}{QA accuracy (\%) $\uparrow$}\\
\cmidrule(lr){2-4}
condition & Qwen & InternVL & LLaVA\\
\midrule
none                 &62.8 &59.0 &70.5\\
knowledge only       &68.2 &71.0 &76.4\\
shuffle image        &63.7 &62.3 &70.8\\
\emph{rag} (matched) &79.3 &76.2 &82.6\\
\bottomrule
\end{tabular*}
\caption{Modality-isolation ablations on the full split ($n{=}28{,}087$). Cross-modal binding, not text leakage: mismatched knowledge (\emph{shuffle image}) returns to \emph{none}, while matched \emph{rag} exceeds both.}
\label{tab:modality}
\end{table}

\section{Source-blinded hallucination}
\label{app:blind}
The judge in Table~\ref{tab:main} sees the injected records in its reference, which can mark a claim \textsc{supported} merely because judge and model saw the same record. Table~\ref{tab:blind} re-runs the judge with the injected knowledge hidden (category-only reference). The \emph{none}$\rightarrow$filtered reduction persists under blinding but shrinks: the 19.28--45.19\% reported with the standard judge becomes 9.58--26.34\%, and absolute rates roughly double once the judge cannot lean on the injected text. Under blinding, full and filtered injection are within $0.3$ points, so the filter's advantage is in conflict robustness and retained accuracy (Appendix~\ref{app:controls}) rather than a lower hallucination rate; we therefore report the blinded reduction in the main text.

\begin{table}[H]
\centering\small
\setlength{\tabcolsep}{4pt}
\begin{tabular*}{\columnwidth}{@{\extracolsep{\fill}}lrrrr@{}}
\toprule
& \multicolumn{3}{c}{blinded halluc.\ (\%) $\downarrow$} & reduction\\
\cmidrule(lr){2-4}
Model & none & full & filt. & blind / raw\\
\midrule
Qwen2.5-VL &11.68 &8.43 &8.60 & $-26.3$ / $-43.8$\\
InternVL3  &10.67 &9.20 &9.45 & $-11.4$ / $-21.2$\\
LLaVA-OV   &13.41 &11.79 &12.13 & $-9.6$ / $-20.3$\\
\bottomrule
\end{tabular*}
\caption{Source-blinded claim-level hallucination (\textsc{contradicted} share, 3{,}000 descriptions/model), judge reference = fMoW category only. Reduction (\%) is \emph{none}$\rightarrow$filtered under the blinded / standard judge.}
\label{tab:blind}
\end{table}

\end{document}